\documentclass[pdflatex,sn-basic]{sn-jnl}

\usepackage{graphicx}%
\usepackage{multirow}%
\usepackage{amsmath,amssymb,amsfonts}%
\usepackage{amsthm}%
\usepackage{mathrsfs}%
\usepackage[title]{appendix}%
\usepackage{xcolor}%
\usepackage{textcomp}%
\usepackage{manyfoot}%
\usepackage{booktabs}%
\usepackage{algorithm}%
\usepackage{algorithmicx}%
\usepackage{algpseudocode}%
\usepackage{listings}%
\usepackage{array}% ragged-right p{} columns in tables
\usepackage{tabularx}% full-width Table 1
\usepackage{pifont}% check / cross marks
\newcommand{\cmark}{\ding{51}}%
\newcommand{\xmark}{\ding{55}}%
\newcommand{\rothead}[1]{\rotatebox{90}{\textbf{#1}}}% vertical column heading (Table 1)

\usepackage[most]{tcolorbox}% prompt box in the appendix
\definecolor{orange}{RGB}{243, 135, 60}
\newtcolorbox{promptbox}[1]{
  colback=white,
  colframe=black,
  colbacktitle=orange!30,
  coltitle=black,
  title=#1,
  fonttitle=\bfseries,
  sharp corners,
  boxrule=0.8pt
}

\usepackage{tikz}%
\definecolor{orcidgreen}{RGB}{166,206,57}
\renewcommand{\orcidlogo}{\,\tikz[baseline=-0.6ex]{%
  \fill[orcidgreen] (0,0) circle (0.75ex);
  \node[text=white,inner sep=0pt,font=\sffamily\bfseries\fontsize{4.5}{4.5}\selectfont] at (0,0) {iD};}}

\begin{document}

\title[LLM-Enhanced Multimodal House Price Prediction]{LLM-Enhanced Multimodal House Price Prediction using Structured, Textual, and Visual Features}

\author[1]{\fnm{Khandokar Md. Rahat} \sur{Hossain}\orcid{https://orcid.org/0009-0006-8003-8587}}\email{kmrahat.buet@gmail.com}

\author[1]{\fnm{Souvik} \sur{Das}\orcid{https://orcid.org/0009-0007-5065-5346}}\email{souvik.das@tamu.edu}
%% \equalcont prints a dagger note after this author; used here for the present address
\equalcont{Present address: Texas A\&M University, College Station, Texas, USA.}

\author[2]{\fnm{Tahsina} \sur{Hashem}\orcid{https://orcid.org/0000-0002-4714-1543}}

\author[2]{\fnm{Mohammed Eunus} \sur{Ali}\orcid{https://orcid.org/0000-0002-0384-7616}}

\author*[1]{\fnm{Tanzima} \sur{Hashem}\orcid{https://orcid.org/0000-0003-1288-5785}}\email{tanzimahashem@cse.buet.ac.bd}

\affil*[1]{\orgname{Bangladesh University of Engineering and Technology}, \orgaddress{\city{Dhaka}, \country{Bangladesh}}}

\affil[2]{\orgname{Monash University}, \orgaddress{\city{Melbourne}, \country{Australia}}}

%%==================================%%
%% Unstructured abstract            %%
%%==================================%%

\abstract{Accurate house price prediction is essential for real estate buyers, sellers, agents, and financial institutions. Traditional methods rely primarily on structured data, e.g., number of bedrooms, bathrooms, parkings, but they overlook rich contextual information embedded in textual descriptions and property photos. Previous deep learning approaches for house price prediction have fused multimodal data but they fall short to extract structured spatio-temporal and contextual features from texts and semantic visual information from images. In this paper, we introduce a novel multimodal framework that addresses these limitations by systematically integrating structured attributes, textual descriptions, and visual images of properties to achieve higher prediction accuracy while maintaining interpretability. We use a large language model to derive domain-specific spatio-temporal and contextual features from text, and an unsupervised vision-language approach to automatically group photos into meaningful classes like indoor, outdoor, mixed, or floorplan, that are commonly present in real-world house listings. These categories are created without manual labeling by aligning image clusters with carefully designed pseudo text prompts. We fuse all features into a unified representation and use an optimized regressor to predict prices. Experiments on a large real-world property dataset demonstrate that our method outperforms traditional methods and existing multimodal baselines. Ablation studies confirm that the novel textual feature extraction and categorized visual embeddings substantially contribute to more accurate and reliable price predictions.}


\maketitle

%% 1. Introduction

\section{Introduction}
\label{sec:intro}

\begin{figure}[htbp]
\centering
\includegraphics[width=\textwidth]{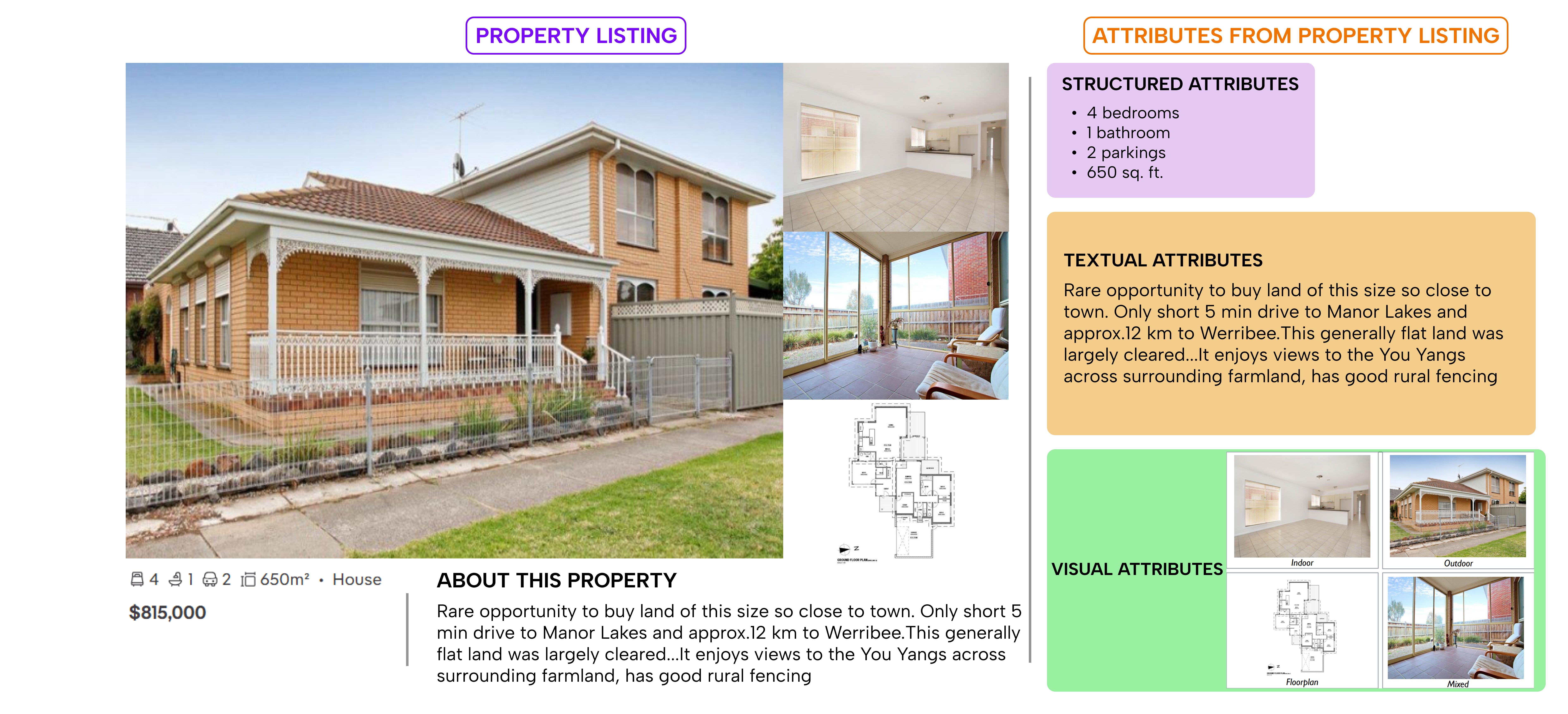}
\caption{Structured, Textual, and Visual Attributes of a Property}
\label{fig:features_breakdown}
\end{figure}

Accurate residential property price prediction remains one of the most challenging and practically important problems in the real estate domain, with significant implications for buyers, sellers, financial institutions, and policymakers. Despite its importance, property valuation is inherently complex, as prices emerge from intricate and non-linear interactions among structured attributes, spatial context, neighborhood characteristics, and qualitative factors that are difficult to quantify. While structured attributes such as the number of bedrooms and bathrooms, floor area, land size, and property type describe the physical characteristics of a property, they often fail to capture many influential factors that shape buyer perception and market value.

For example, properties with similar size and location may still exhibit substantial price differences due to factors such as aesthetic appeal, interior quality, scenic surroundings, accessibility to important facilities, or the perceived attractiveness of the neighborhood. These factors are rarely available in structured form, making them difficult to incorporate into conventional machine learning models. However, modern online property listings naturally contain much of this hidden information through textual descriptions and listing images. A property description may highlight that a house is ``only a few minutes from the town center" or ``offers stunning park views," while images may reveal a beautifully landscaped garden, renovated interiors, or bright modern living spaces. Such textual and visual cues strongly influence buyer preferences and willingness to pay. Figure~\ref{fig:features_breakdown} shows the heterogeneous structured, textual, and visual attributes available for a property.

Therefore, effective house price prediction requires a multimodal understanding that integrates structured attributes with semantic information extracted from text and images. Combining these heterogeneous modalities presents significant challenges due to their diverse characteristics and complex non-linear relationships. Beyond predictive accuracy, interpretability is equally important, as stakeholders need to understand why a property receives a particular valuation. To address these challenges, we propose a framework that extracts semantically meaningful and human-interpretable representations from each modality before performing transparent multimodal fusion, enabling both improved predictive performance and clearer explanations of the factors driving property valuation.

Early work on house price prediction relied on hedonic pricing models \citep{owusu2011review,can1992specification}, which represent property value as a function of carefully selected structural attributes. Subsequent studies incorporated spatial dependence and neighborhood effects using spatial econometric techniques \citep{bourassa2003housing,bourassa2007spatial,anselin1988spatial,lesage2009introduction,anselin2009spatial}, demonstrating the importance of explicitly modeling location-related interactions. While effective, these approaches require extensive domain expertise and manual feature engineering, limiting their scalability across different housing markets.

With the availability of large-scale real estate data, machine learning methods have become prominent alternatives \citep{wang2014real,chen2017house,piao2019housing,zhao2019deep,xiong2019improve}. Ensemble tree-based models such as XGBoost and LightGBM \citep{chen2016xgboost,ke2017lightgbm} have shown strong performance on structured tabular data, while recent work has explored geo-spatial embeddings and network-based representations \citep{das2021boosting,karamanou2024explainable}. However, approaches relying solely on structured data often overlook important latent factors that influence buyer perception and willingness to pay, such as interior quality, renovation status, and aesthetic appeal.

Recent studies have explored multimodal learning approaches that combine structured attributes with listing photos, street-level, map, or satellite imagery, and textual descriptions for house price prediction \citep{poursaeed2018vision,law2019take,bin2019attention,nouriani2022vision,hasan2024multi}. While these methods have demonstrated the value of incorporating heterogeneous data sources, many rely on end-to-end deep architectures that directly fuse high-dimensional text and image embeddings, and those that extract interpretable image features typically depend on classifiers trained on labeled photos. Such representations often fail to explicitly capture domain-relevant factors that drive housing prices. For example, textual descriptions frequently contain valuable contextual information regarding proximity to amenities, neighborhood characteristics, renovation status; while property images reveal cues related to interior quality, curb appeal. When represented only as generic embeddings without explicit semantic understanding, these signals become less meaningful to analyze and interpret as factors for price prediction.

In this paper, we propose a novel, multimodal pipeline for house price prediction that effectively addresses key limitations of existing approaches. Our main contributions are summarized as follows:

\begin{itemize}
\item We propose a modular multimodal framework that integrates structured attributes, contextual signals extracted from textual descriptions, and semantic information derived from unlabeled property images, avoiding opaque end-to-end fusion architectures.

\item We introduce a structured extraction method based on large language models (LLM) that converts free-form real estate descriptions into key spatio-temporal and contextual features using prompt engineering.

\item We design an unsupervised visual representation pipeline that leverages the vision-language model (VLM) and pseudo-prompt engineering to categorize unlabeled property images into semantically meaningful groups, enabling category-aware visual feature construction without manual annotation.

\item Extensive experiments and ablation studies demonstrate that both LLM-derived contextual features and category-aware visual semantics provide consistent and complementary improvements over traditional structured and spatial feature-based approaches.
\end{itemize}

\begin{table}[htbp]
\caption{Comparison of representative house price prediction approaches.}
\label{tab:related_summary}
\begin{tabular}{@{}lccccc l@{}}
\toprule
& & \multicolumn{2}{c}{\textbf{Textual}}
& \multicolumn{2}{c}{\textbf{Visual}} & \\
\cmidrule(lr){3-4} \cmidrule(lr){5-6}
\textbf{Approach} & \rothead{Structured}
& \rothead{Spatio-temporal} & \rothead{Contextual}
& \rothead{Non-categorized} & \rothead{Categorized} & \textbf{Method} \\
\midrule
Hedonic        & \cmark & \xmark & \xmark & \xmark & \xmark & Linear Hedonic regression \\
Traditional Machine Learning & \cmark & \xmark & \xmark & \xmark & \xmark & SVM, RF, XGBoost \\
Geo-spatial     & \cmark & \cmark & \xmark & \xmark & \xmark & Spatial Autoregressive Estimator \\
NLP-based      & \cmark & \xmark & \cmark & \xmark & \xmark & LLM \\
Graph-based    & \cmark & \cmark & \xmark & \xmark & \xmark & GCN, GNN \\
Multimodal Deep Learning    & \cmark & \xmark & \cmark & \cmark & \xmark & LLM, CNN, Attention Fusion \\
\midrule
\textbf{Ours} & \textbf{\cmark} & \textbf{\cmark} & \textbf{\cmark}
& \textbf{\cmark} & \textbf{\cmark} & \textbf{LLM + VLM + Regressor} \\
\botrule
\end{tabular}
\footnotetext{SVM: support vector machine; RF: random forest; NLP: natural language processing; LLM: large language model; GCN: graph convolutional network; GNN: graph neural network; CNN: convolutional neural network; VLM: vision-language model.}
\end{table}

%% 2. Related Work

\section{Related Work}
\label{sec:related}

The growing importance of accurate residential property price prediction, driven by the rapid expansion of online real estate platforms, increasing housing market complexity, and the widespread availability of large-scale property data, has made property price prediction a highly active area of research over the past two decades. As the limitations of traditional prediction techniques became increasingly evident, researchers have continuously explored more sophisticated computational approaches capable of modeling the complex relationships among diverse factors influencing property values. Consequently, the field has evolved from conventional statistical and econometric models to advanced machine learning, deep learning, graph-based, and multimodal frameworks that integrate structured property attributes with textual descriptions, geospatial information, and visual content. Each successive generation of methods has sought to improve predictive accuracy by capturing increasingly rich and heterogeneous sources of information. This continuous evolution has not only advanced the state of the art in automated property valuation but has also highlighted the importance of leveraging complementary data modalities to better represent real-world market dynamics.

The section is organized as follows. Section \ref{sec:hedonic} gives an overview of the hedonic models for house price predictions and their shortcomings. Section \ref{sec:traditional-ml} takes through the traditional machine learning models and how they handled certain features. Section \ref{sec:geospatial} emphasizes on the location based models and their key innovations. Section \ref{sec:nlp-based} discusses how natural language processing (NLP) models were used in extraction of contextual features from textual descriptions of raw property listings. Finally, section \ref{sec:multimodal} enlightens how multimodal approaches started to shed light on the house price prediction domain and how they became quite successful than all the previous approaches.

Table~\ref{tab:related_summary} summarizes representative house price prediction approaches across structured, textual, and visual modalities, highlighting the key differences between prior works and our proposed framework.

\subsection{Hedonic Models}
\label{sec:hedonic}

\subsubsection{Classical Hedonic Pricing Theory}

Early research was based on hedonic pricing theory, which models property value as a function of observable structural and neighborhood characteristics. \citet{can1992specification} proposed traditional and spatial autoregressive hedonic models, emphasizing that prices are determined by a bundle of structural, neighborhood, and location attributes and highlighting spatial dependence. \citet{owusu2011review} reviewed functional forms (linear, semi-log, log-log, Box-Cox) and estimation techniques, rooted in Rosen's~\citeyearpar{rosen1974hedonic} implicit market theory, and identified challenges such as omitted variable bias, spatial autocorrelation, and functional form misspecification.

Additional foundational works include \citet{sheppard1999hedonic} who surveyed hedonic analysis in housing markets, and \citet{malpezzi2003hedonic} who discussed practical applications and limitations. \citet{helbich2013spatial} reviewed spatial heterogeneity in hedonic models. Further theoretical and empirical advancements in hedonic modeling were contributed by \citet{freeman1979hedonic}, who explored willingness to pay for public goods, \citet{epple1987hedonic} on demand and supply estimation, \citet{quigley1984nonlinear} on nonlinear budget lines, \citet{palmquist1984hedonic} on demand estimation for housing characteristics. \citet{osland2008spatial} and \citet{jud2002spatial} provided important reviews on spatial variation in hedonic models. These studies collectively established the robustness of hedonic frameworks but also revealed their limitations in handling unstructured data and complex interactions.

\subsubsection{Machine Learning Assisted Hedonic Models}

More recent work has continued to probe the linear assumptions underlying hedonic price models and to benchmark them directly against machine learning alternatives. \citet{an2025toward} compared ordinary-least-squares and spatial-lag hedonic specifications against random forest and deep neural network models on a large South Korean condominium dataset, and used SHapley Additive exPlanations (SHAP) \citep{louhichi2023shapley} to recover the direction, not just the magnitude, of each hedonic variable's contribution to price, something Gini importance and permutation importance cannot provide on their own. Their results showed that a random forest model paired with SHAP could match the interpretive transparency of a classical hedonic regression while substantially improving predictive accuracy, and that the sign of SHAP contributions largely agreed with the sign of the corresponding hedonic regression coefficients, lending statistical support to the machine-learning explanations. In a similar spirit, \citet{deppner2023boosting} examined U.S. commercial real estate appraisals using NCREIF Property Index data spanning 1997-2021 and found that gradient-boosting models could shrink the deviation between appraised market values and eventual transaction prices beyond what hedonic-style valuation models achieve, though the gains were less pronounced than in the residential sector because commercial data are sparser and less uniform. \citet{embaye2021predicting} took the comparison to a data-scarce setting, evaluating hedonic ordinary least square (OLS) regression against ridge, LASSO, tree, random forest, bagging, and boosting models for predicting rental values from household survey data in Tanzania, Uganda, and Malawi, while also correcting for spatial autocorrelation using spatial error and spatial autoregressive terms; they found that machine learning approaches offered modest but consistent accuracy gains even with limited, survey-based samples. \citet{lorenz2023interpretable} pushed the interpretability debate further by applying model-agnostic explanation methods to commercial real estate machine learning models, arguing that explainability techniques can recover economically meaningful, hedonic-style relationships from otherwise opaque models, thereby narrowing the perceived gap between classical hedonic theory and modern predictive methods.

\subsubsection{Spatial and Semiparametric Hedonic Models}

Beyond comparisons with machine learning, researchers have also continued to refine the hedonic model's own statistical machinery to better handle heterogeneity and nonlinearity without abandoning its interpretable, parametric structure. \citet{schirripa2024spatial} proposed a spatial semiparametric M-quantile regression for hedonic price modelling, combining univariate and bivariate spline components with M-quantile estimation to simultaneously capture nonlinear amenity effects (such as the value of cultural or environmental features) and spatial price trends, while remaining robust to outliers and market heterogeneity across different segments of the price distribution. This line of work illustrates that the hedonic tradition has not been static in the face of machine learning competition: rather than being displaced outright, it has absorbed some of the same flexibility-oriented motivations like nonlinearity, spatial dependence, distributional heterogeneity, into extensions that preserve the coefficient-level interpretability that makes hedonic models attractive to practitioners and policymakers.

Overall, hedonic models provide strong economic interpretability and have served as the foundation of residential valuation studies. However, they rely heavily on manually specified functional forms and often struggle to capture complex nonlinear interactions and latent factors influencing buyer perception, motivating the adoption of more flexible machine learning approaches.

\subsection{Traditional Machine Learning Approaches}
\label{sec:traditional-ml}

\subsubsection{Conventional Machine Learning Models}

Traditional machine learning (ML) methods marked a significant shift from rigid econometric models by offering greater flexibility in capturing non-linear patterns. The increasing availability of large-scale real estate datasets has accelerated the use of machine learning techniques for property valuation. Methods including support vector machines, ensemble regressors, boosting frameworks, convolutional neural networks (CNNs), and long short-term memory networks (LSTMs) have demonstrated substantial improvements over traditional econometric models by capturing nonlinear relationships among housing attributes \citep{wang2014real,zhao2019deep,chen2016xgboost,ke2017lightgbm,xiong2019improve,serigne2017stacked,piao2019housing,chen2017house}.

\subsubsection{Deep Learning Models}

\citet{wang2014real} proposed a house price prediction framework that combined Support Vector Machines (SVM) with Particle Swarm Optimization (PSO) to optimize the model parameters automatically. Their optimization strategy significantly improved predictive accuracy by enhancing the generalization capability of the SVM model and demonstrated the effectiveness of swarm intelligence for real estate valuation. \citet{chen2017house} introduced a Long Short-Term Memory (LSTM)-based approach to model the temporal dependencies present in housing market data. By exploiting sequential information from historical transactions, their model captured temporal price trends more effectively than conventional machine learning techniques. \citet{piao2019housing} employed Convolutional Neural Networks (CNNs) to learn complex feature representations from housing-related data. Their study demonstrated that CNNs could effectively extract local spatial patterns and nonlinear relationships among property characteristics, resulting in improved house price estimation. \citet{zhao2019deep} proposed a hybrid framework that integrated deep neural networks with XGBoost to leverage the complementary strengths of representation learning and gradient boosting. The combination enabled the model to capture both high-level feature interactions and nonlinear relationships, leading to superior predictive performance compared with individual learning algorithms. \citet{xiong2019improve} developed a stacked generalization ensemble model that combined the predictions of multiple base learners through a meta-learning strategy. Their ensemble framework effectively reduced prediction errors and improved robustness by exploiting the diversity of different machine learning models.

Collectively, these approaches leveraged ensemble methods such as XGBoost, LightGBM, and deep learning architectures to better model complex interactions among housing attributes.

\subsubsection{Hybrid and Ensemble Learning}

Some other notable contributions included hybrid and ensemble learning, such as \citet{de2011ames}, who popularized the Ames housing dataset for regression benchmarking, and various ensemble-based studies \citep{serigne2017stacked, zhao2019deep}. More recent works have explored hybrid models combining traditional ML with deep learning components \citep{li2022imbalanced, sellam2025boosting}.

A further line of work has systematically benchmarked machine learning models directly against each other and against hedonic baselines on residential transaction data. \citet{ho2021predicting} compared support vector regression, random forest, and gradient boosting models for predicting apartment prices in Hong Kong, finding that ensemble tree-based methods consistently outperformed both single-model machine learning approaches and traditional hedonic regressions, particularly for higher-value properties where nonlinearities in the price-attribute relationship are more pronounced. Their results reinforced a broader pattern in the literature: tree-based ensembles tend to offer the best accuracy-to-interpretability trade-off among purely tabular, non-spatial, non-textual modeling approaches, which is part of why they continue to serve as strong baselines even as researchers move toward incorporating spatial, textual, and visual information.

\subsubsection{Loss Function Optimization-Focused}

The choice of loss function within gradient-boosted tree models has also received dedicated attention, since the standard squared-error objective implicitly treats over- and under-prediction symmetrically even though the practical cost of misvaluing a property is often asymmetric, like in mortgage underwriting or automated valuation for taxation. \citet{hjort2022house} systematically compared XGBoost models trained under different loss functions for house price prediction, showing that the choice of loss function meaningfully affects both point-prediction accuracy and the shape of the resulting error distribution, and arguing that practitioners should select a loss function aligned with the specific downstream decision the valuation is meant to support rather than defaulting to squared error. This kind of methodological refinement of an otherwise well-established ensemble technique illustrates that gains in the traditional machine learning literature have come not only from adopting new model families, but also from more carefully tuning the training objectives of established ones.

Despite their predictive success, most approaches remain focused on structured tabular data and largely ignored the rich semantic information available in textual descriptions and property images.

\subsection{Geo-spatial Approaches}
\label{sec:geospatial}

\subsubsection{Spatial Econometric Models}

Location is a primary determinant of property value. Spatial econometric techniques explicitly model neighborhood effects and spatial dependence \citep{anselin1988spatial,lesage2009introduction,anselin2009spatial}, while recent approaches employ geo-spatial embeddings, graph representations, and streetscape features \citep{luo2025deciphering,han2025effects}.

\subsubsection{Graph-based Spatial Learning}

\citet{das2021boosting} introduced Geo-Spatial Network Embedding (GSNE) using graph neural networks (GNNs). Additional studies include \citet{bourassa2003housing,bourassa2007spatial} on housing submarkets, and recent GNN applications \citep{riveros2024scalable}. The rapid uptake of GNNs for spatial house-price modeling has itself become a subject of scrutiny. \citet{karamanou2024explainable} applied three representative GNN architectures, namely ChebNet, GCN, and GraphSAGE, to open statistical knowledge graphs built from Scottish government data, showing that graph-based spatial dependency modeling could outperform tabular baselines such as XGBoost and a simple multi-layer perceptron, while also using local and global explainability techniques to identify which socio-economic indicators drove each prediction. \citet{mostofi2023multiedge} proposed a Multiedge Graph Convolutional Network that encodes distance, zoning, and infrastructure relationships as multiple edge types between properties, allowing the model to capture more nuanced spatial context than a single homogeneous adjacency graph, though at the cost of sensitivity to missing or conflicting geospatial data. However, not all comparative evidence favors graph-based spatial modeling. In a large benchmarking study spanning six diverse housing datasets and multiple graph construction strategies, \citet{geerts2025graph} found that GNNs were generally unable to outperform well-tuned tree-based models such as LightGBM and CatBoost for house price prediction, raising questions about whether the added architectural complexity and computational cost of graph-based spatial encodings are justified relative to simpler spatial features engineered directly into tabular models. This tension between the theoretical appeal of graph-structured spatial reasoning and its inconsistent empirical payoff remains an open methodological question in the geo-spatial house price prediction literature.

\subsubsection{Streetscape and Visual Spatial Features}

\citet{fu2019street} combined house records with street-view panoramas, using deep segmentation to compute greenery, sky, and building view indexes and incorporating them into a hedonic model. Their results showed that perceptual scene qualities carry economically meaningful information orthogonal to point-of-interest (POI) distance and network features.

Although these methods improve spatial representation, they generally do not exploit the full semantic richness of textual descriptions and property images.

\subsection{NLP-based Approaches}
\label{sec:nlp-based}

\subsubsection{Transformer-based Text Representation}

Property descriptions often contain valuable information regarding renovation status, accessibility, neighborhood quality, scenic views, lifestyle appeal, and other qualitative factors that influence valuation. Transformer models such as BERT and its variants introduced contextual representations \citep{reimers2019sentence,sanh2019distilbert,devlin2019bert,liu2019roberta}. Large language models further enabled structured extraction from free-form text \citep{touvron2023llama,brown2020language,raffel2020exploring}. Most existing NLP approaches in real estate, however, still use textual models primarily as embedding generators rather than for explicit domain-specific feature extraction.

\subsubsection{LLM-based Structured Feature Extraction}

\citet{zhang2023describe} showed that features mined from free-text listings (renovation, condition, amenities) carry predictive signal not fully captured by structured attributes. Subsequent work has used instruction-following LLMs to extract quantifiable variables such as renovation status, proximity to green space, or view quality, aiming to preserve hedonic-style interpretability while retaining the flexibility of language models.

\subsubsection{Large Language Models for Property Valuation}

\citet{geerts2025llm} evaluated leading LLMs on international housing datasets using zero-shot, few-shot, and hybrid in-context learning. Carefully selected examples improved estimates, and LLM explanations largely agreed with feature importances of strong ML models. However, LLMs still lag dedicated ML models on pure accuracy and tend to be overconfident, suggesting they are better suited to interpretable feature extraction than to serving as the primary valuation engine.

\subsection{Multimodal House Price Prediction}
\label{sec:multimodal}

\subsubsection{Vision-based Property Valuation}

Vision-based methods using property photographs, satellite imagery, and street-view data have shown that visual cues provide complementary information \citep{poursaeed2018vision,law2019take,nouriani2022vision}. \citet{poursaeed2018vision} explored vision-based estimation without image categorization. \citet{law2019take} used street-view and satellite images.

\subsubsection{Multimodal Fusion Architectures}

Attention-based and deep multimodal frameworks jointly learn representations across modalities \citep{bin2019attention,hasan2024multi,zhao2024house}. \citet{bin2019attention} proposed attention-based fusion. \citet{nouriani2022vision} incorporated interior, exterior, and satellite images. \citet{hasan2024multi} combined raw features, geospatial context, text, and images. Models such as CLIP (Contrastive Language-Image Pre-training) and LLMs have further strengthened multimodal representation learning \citep{radford2021learning,touvron2023llama}. Additional studies include \citet{zhao2024house}, \citet{huang2025multimodalsurvey}, and \citet{li2022imbalanced}.

\subsubsection{Image-based Feature Extraction and Categorization}

\citet{yousif2023real} constructed the REPD-3000 dataset with room-level images and showed that CNN-derived visual features add predictive value beyond a single whole-house embedding. \citet{deng2024diffusion} used CLIP-based hierarchical classification with diffusion-based augmentation to label luxury tier and room type.

\subsubsection{Multimodal Learning Surveys}

\citet{huang2025multimodalsurvey} surveyed multimodal machine learning for real estate appraisal, classifying modalities (structured, textual, visual, geospatial) and identifying a persistent gap in quantifying each modality's individual contribution to fused predictions---the opacity problem that motivates our transparent, feature-centric design.

\subsubsection{Explainability in House Price Prediction}

Interpretability has become increasingly important. \citet{louhichi2023shapley} explored Shapley values; \citet{ribeiro2016lime} introduced LIME; \citet{lundberg2017unified} proposed a unified approach; \citet{molnar2020interpretable} provided a comprehensive guide.

\subsection{Research Gaps}

While these multimodal methods demonstrated the value of heterogeneous data sources, most rely on end-to-end neural fusion of raw or high-dimensional embeddings. This limits interpretability and makes it difficult to isolate the contribution of individual modalities or extract explicit domain-relevant signals such as proximity advantages, interior quality, curb appeal, or other qualitative factors.

In contrast, our framework adopts a transparent, feature-centric design. We extract structured contextual features from text using LLM-based prompting and derive category-aware visual features through unsupervised pseudo-prompt engineering with CLIP. This modular approach addresses the opacity and limited semantic understanding of previous multimodal methods while enabling better interpretability and performance.

%% 3. Methodology

\begin{figure}[htbp]
\centering
\includegraphics[width=0.9\textwidth]{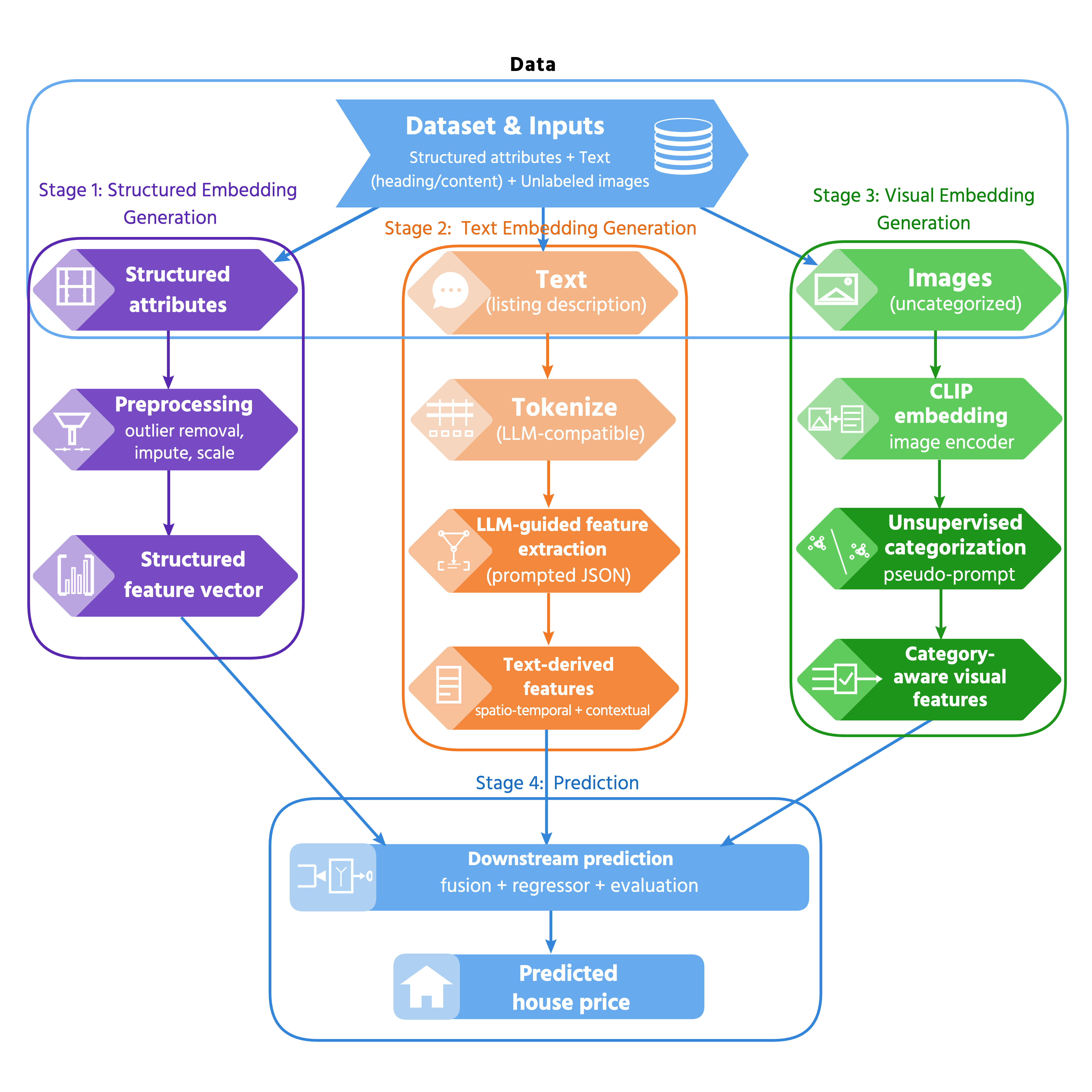}
\caption{End-to-end Pipeline for LLM/VLM-Enhanced Multimodal House Price Prediction}
\label{fig:pipeline}
\end{figure}

\section{Methodology}
Our proposed framework is an end-to-end multimodal pipeline designed to predict residential property prices by systematically integrating three complementary data modalities: structured numerical and categorical property attributes, unstructured textual property descriptions, and property images. The central motivation here is that property value emerges from complex, non-linear interactions across quantitative measurements, textual context, and visual appeals (Figure \ref{fig:features_breakdown}).

The overall pipeline is organized into four main stages: (1) structured embedding generation, (2) textual embedding generation, (3) visual embedding generation, and (4) multimodal fusion. Figure \ref{fig:pipeline} illustrates the end-to-end pipeline of the proposed LLM/VLM-enhanced multimodal framework for house price prediction. The model ingests heterogeneous inputs consisting of structured property attributes, free-form listing text, and unlabeled images. Structured attributes are preprocessed and encoded into numerical feature vectors. Textual descriptions are tokenized and processed using a prompt-engineered LLM to extract spatio-temporal and contextual features in a structured format. In parallel, images are embedded using a pretrained CLIP encoder, followed by unsupervised categorization via pseudo-prompt engineering to derive category-aware visual features. The resulting modality-specific representations are then combined within a downstream prediction module to estimate house prices. Darker blocks denote components introduced in this work, while lighter blocks correspond to standard or reused processing steps.

\subsection{Structured Embedding Generation}

Structured attributes form the quantitative basis for property valuation. They include continuous measurements (e.g., number of bedrooms, distance to transport hubs) and discrete categorical indicators (e.g., property type, locality).

\subsubsection{Data Preprocessing}
\label{subsec:structured_preprocessing}

\textbf{Initial Data Cleaning and Filtering.} Missing numerical fields underwent initial imputations using median values, while categorical fields were updated by the mode or default category to maintain consistency. To ensure temporal consistency and enable fair comparison across years, historical sale prices were adjusted forward to a common reference year.

Temporal consistency for sale prices was adjusted by applying a compound annual growth factor derived from a reliable inflation or housing price index. The adjustment formula is:

\begin{equation}
P_{\text{adjusted}} = P_{\text{original}} \times (1 + r)^{t_{\text{ref}} - t_{\text{sale}}}
\end{equation}

where:
\begin{itemize}
    \item $P_{\text{adjusted}}$ is the inflation-adjusted sale price
    \item $P_{\text{original}}$ is the nominal transaction price recorded at the time of sale
    \item $r$ is the annual inflation rate
    \item $t_{\text{ref}}$ is the reference year
    \item $t_{\text{sale}}$ is the year of the original transaction
\end{itemize}

In our implementation, we used a conservative constant annual growth rate of 6\% consistent with long-term average housing inflation in Australia at the time of writing this paper, applied backward from the reference year to align all prices to current-equivalent values.

\subsubsection{Structured Feature Vector Construction}

A carefully designed set of feature transformations were applied to the structured attributes to better capture the real-world factors that influence property prices.

First, we enhanced the location-related features. Raw distance measurements (e.g., distance to the nearest train station) were converted from meters to kilometers and then squared. This transformation reflects the intuition that the benefit of being close to amenities decreases non-linearly.

Second, we created interaction features to model combined effects between attributes. We systematically generated candidate interaction terms by multiplying pairs of numerical features. From the resulting candidates, we retained only those interactions that improved predictive performance on the validation set, measured by a reduction in root mean squared error (RMSE). For example, the product of the number of bedrooms and local rental prices was retained as the feature \texttt{rent\_bedroom\_interaction}. Similarly, the product of school ranking and proximity to schools was kept. These multiplicative terms allow the model to learn synergistic relationships that simple additive features cannot capture.

Third, we incorporated neighborhood context by calculating summary statistics (mean and standard deviation of sold prices) for each locality and merging them back into the dataset. This helps the model understand whether a property is located in a generally expensive, affordable, or highly variable neighborhood.

\subsection{Textual Embedding Generation}

Property listings contain rich unstructured textual descriptions that often include important details about location advantages, property condition, surrounding environment, seller intent etc. To unveil this information, we employed an LLM for zero-shot structured feature extraction from textual descriptions.

\subsubsection{LLM-Guided Feature Extraction}

\begin{figure}[htbp]
\centering
\includegraphics[width=\textwidth]{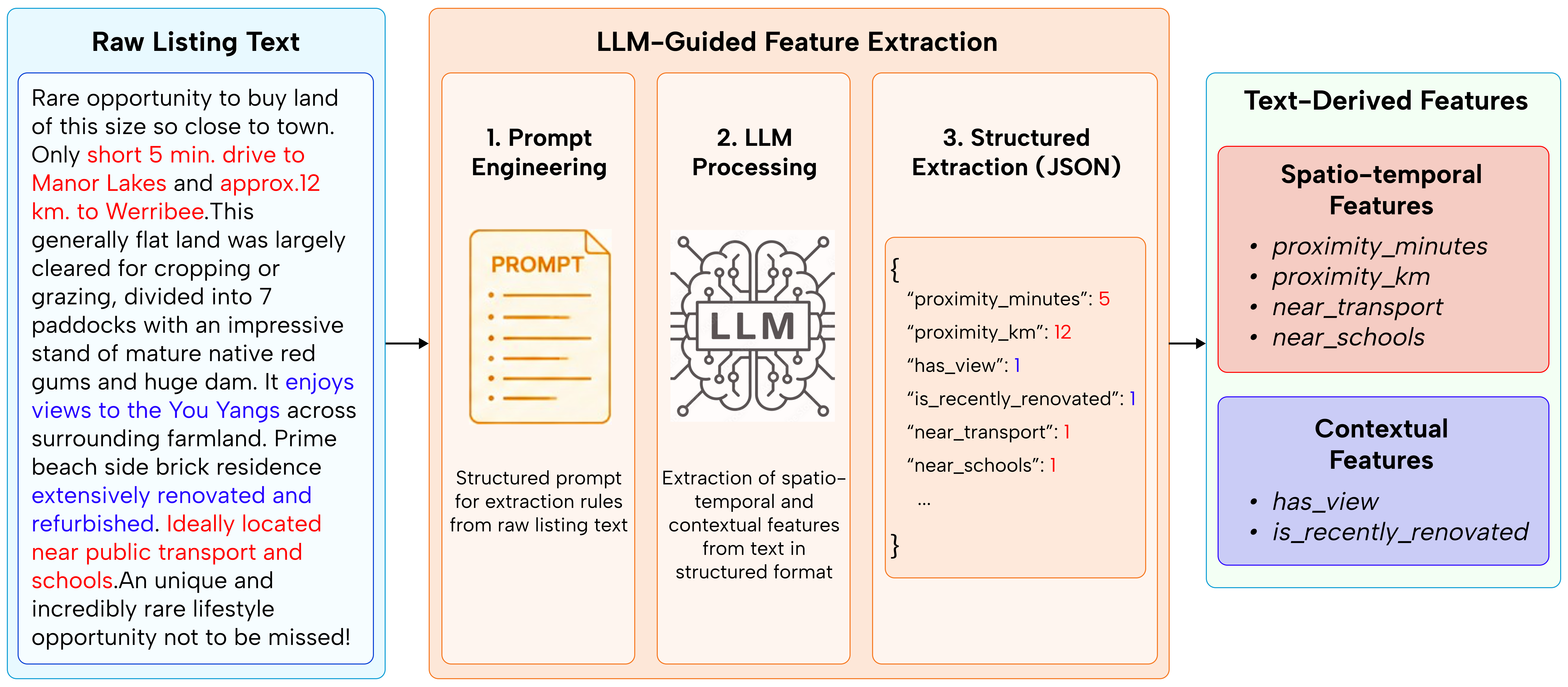}
\caption{Spatio-Temporal and Contextual Feature Extraction using LLM Prompt}
\label{fig:llm_extraction}
\end{figure}

We have developed a generalized prompt engineering approach that can flexibly extract a wide range of spatio-temporal and contextual features from any real estate description. The prompt is designed to identify relevant attributes such as proximity to amenities, neighborhood characteristics, property condition, renovation status, market urgency, perceived quality etc. which are commonly important across different property markets.

We carefully crafted the spatio-temporal and contextual features based on common value drivers observed in real estate markets and our dataset analysis. These features are selected in a way that frequently appear in property descriptions and have strong influence on buyer decisions and final sale prices. We instructed the LLM to look for indicators related to those features. For example:
\begin{itemize}
    \item Spatio-temporal: location relative to amenities and landmarks (shops, public
      transport, schools, parks, water, city, employment hubs), proximity and
      accessibility, orientation, time in market context, etc.
    \item Contextual: condition, age, build quality, prestige, ambiance, layout and
      space, indoor/outdoor features, suitability, seller intent, etc.
\end{itemize}

An example prompt template used for LLM is presented in \emph{``Spatio-temporal \& Contextual Features Extraction Prompt"} block.

\begin{promptbox}{Spatio-temporal \& Contextual Features Extraction Prompt}
\label{prompt:feature-selection}

You are an expert real-estate analyst. You read a property listing description from any location, market, or writing style and identify the KEY spatio-temporal and contextual features that influence the property's value.\\

You are not being given a predefined list of features. Extract which attributes are salient in each description from the following two sets of features:\\
- Spatio-temporal: location relative to amenities and landmarks (shops, public transport, schools, parks, water, city, employment hubs), proximity and accessibility, orientation, time in market context, etc.\\
- Contextual: condition, age, build quality, prestige, ambiance, layout and space, indoor/outdoor features, suitability, seller intent, etc.\\

Rules for the JSON output:\\
- Output ONLY a single flat JSON object. Keys are concise canonical feature names. Values are numbers.\\
- Use 1 for a present or true qualitative attribute. Use an actual number for a quantity (e.g. minutes, kilometers, counts).\\
- Reuse the SAME canonical name for the same concept across different listings (e.g. always ``near\_station", never ``close\_to\_station" in one and ``station\_nearby" in another) so features are compatible.\\
- Prefer widely-applicable names over one-off phrases, and return at most about {max\_feats} features.
\end{promptbox}

For modern LLMs, even open-ended prompts with some textual descriptions might be able to capture such features. However, instead of using completely open-ended prompts, we used a structured zero-shot prompt that explicitly instructs the LLM to return only the most recurring and impactful features in a specified format.
An example textual description, along with its extracted features highlighted using two distinct colors for each feature type, is shown in Figure~\ref{fig:llm_extraction}.

\subsubsection{Text-derived Feature Extraction}

The text-derived feature extraction finally produces key spatio-temporal and contextual features. By systematically extracting both quantitative proximity signals and qualitative perceptual cues from unstructured text, the model gains access to information that traditional structured-only approaches cannot capture, thus substantially enhancing predictive power of our model's contextual analysis.

\subsection{Visual Embedding Generation}
\label{subsec:visual_embedding_generation}

\begin{figure}[htbp]
\centering
\includegraphics[width=\textwidth]{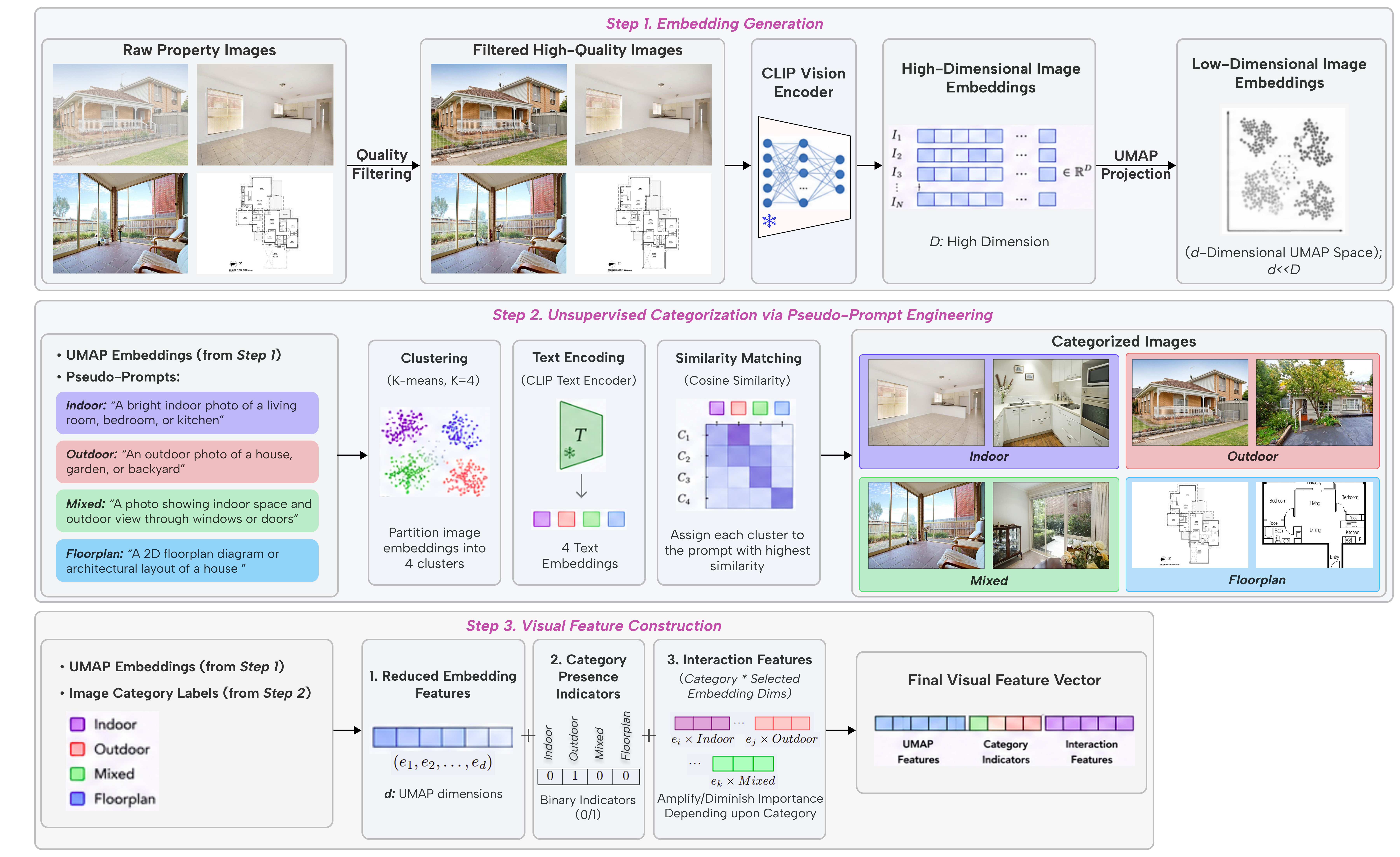}
\caption{Visual Embedding Generation Pipeline}
\label{fig:visual_embed_gen}
\end{figure}

Property images provide direct perceptual evidence of interior condition, exterior appeal, and overall atmospheric characteristics that substantially influence buyer perception and willingness to pay. However, raw high-dimensional images are semantically ambiguous and prone to overfitting due to the curse of dimensionality. Without semantic grouping, image embeddings remain low-level visual descriptors that lack explicit alignment with buyer-oriented attributes such as interior quality, curb appeal, and indoor--outdoor flow. This limitation reduces interpretability and weakens the model's ability to capture modality-specific price signals.

To address this issue, we propose an unsupervised visual embedding generation pipeline that discovers natural visual groupings and assigns semantically meaningful categories to property images. This categorization process transforms raw embeddings into compact and interpretable features that explicitly represent buyer priorities. Specifically, we categorize property images into four meaningful classes: indoor, outdoor, mixed, and floorplan. This step bridges the gap between low-level pixel patterns and high-level real estate value drivers, enabling the downstream model to leverage visual information more effectively.

Figure \ref{fig:visual_embed_gen} illustrates the architecture of the proposed visual embedding generation pipeline. The framework consists of three steps: (1) Embedding Generation, (2) Unsupervised Categorization via Pseudo-Prompt Engineering, and (3) Visual Feature Construction. The following three subsections describe each step in detail.

\subsubsection{Embedding Generation}
\label{appix:embed_generation}
The embedding generation process involves the following steps:

\textbf{Image Quality Filtering.} Real estate listing images vary significantly in quality, from high-resolution professional photographs to low-quality mobile snapshots. Low-resolution, blurry, or poorly lit images introduce noise into visual representation learning. Therefore, we applied quality filtering based on two fundamental criteria: minimum resolution and image sharpness. Images below a predefined resolution threshold were removed, while sharpness was assessed using the variance of the Laplacian operator on grayscale images. Higher variance values indicate sharper images with clearer edges and finer details, whereas lower variance values indicate blur or insufficient detail.

\textbf{CLIP Embedding Generation.} Retained images were encoded using CLIP, a vision-language foundation model pre-trained on a large corpus of image--text pairs. CLIP learns a joint embedding space in which images and their corresponding textual descriptions are mapped close together, while dissimilar pairs are pushed apart. This is achieved through a symmetric contrastive loss function that maximizes the cosine similarity between matching image--text pairs and minimizes it for non-matching pairs within a batch:
\begin{equation}
\mathcal{L} = -\frac{1}{N}\sum_{i=1}^{N} \left[ \log \frac{\exp(\text{sim}(i,t_i)/\tau)}{\sum_{j=1}^{N} \exp(\text{sim}(i,t_j)/\tau)} + \log \frac{\exp(\text{sim}(t_i,i)/\tau)}{\sum_{j=1}^{N} \exp(\text{sim}(t_j,i)/\tau)} \right]
\end{equation}
where $\text{sim}(\cdot,\cdot)$ is cosine similarity (Eq. \ref{cos:sim}), $\tau$ is a temperature parameter scaling the logits, and $N$ is the batch size.

The contrastive loss encourages the model to learn representations where semantically related images and texts are close in the embedding space. This pre-training objective enables CLIP to align arbitrary images with natural-language descriptions. In our pipeline, each preprocessed image was passed through CLIP's vision encoder, which produced a high-dimensional representation that captures rich semantic information about objects, scenes, lighting, composition, and style. By leveraging CLIP's joint representation space, we ensured that the extracted visual features were semantically meaningful and directly comparable to textual prompts, enabling effective zero-shot categorization.

\textbf{Dimensionality Reduction.}
The high-dimensional CLIP embeddings were projected into a lower-dimensional space using a manifold learning technique. The manifold hypothesis posits that high-dimensional data lie on or near a low-dimensional manifold embedded within the ambient space. In the context of property images, semantically similar photos are located close to each other on this manifold, while dissimilar photos (e.g., indoor vs. outdoor scenes) are farther apart. This projection allows us to preserve these intrinsic relationships while mitigating the curse of dimensionality, which can degrade distance measures, amplify noise, and increase computational complexity.

We employed UMAP (Uniform Manifold Approximation and Projection) \citep{mcinnes2018umap}, a modern manifold learning method that preserves both local and global structure in high-dimensional data. Compared to principal component analysis (PCA) \citep{jolliffe2016pca} and t-distributed stochastic neighbor embedding (t-SNE) \citep{van2008visualizing}, UMAP offers better scalability and more stable embeddings while balancing local fidelity and global topology. In our pipeline, it reduces noise in CLIP embeddings, brings semantically similar images closer together, and enables more effective clustering. The resulting low-dimensional representations retain the semantic richness learned by CLIP while mitigating the curse of dimensionality, thereby improving the quality of visual features for price prediction.

\subsubsection{Unsupervised Categorization via Pseudo-Prompt Engineering}

To categorize the property images into four meaningful semantic classes that are most prevalent in property images, i.e., indoor, outdoor, mixed, and floorplan; without manual labeling or any supervised training, we first applied K-means clustering (with K = 4) on the low-dimensional image embeddings obtained from the UMAP reduction, that partitioned the images into four clusters based on visual similarity in the CLIP embedding space. Next, we employed pseudo-prompt engineering to assign meaningful labels to each image cluster. Carefully crafted natural language pseudo-prompts were designed to serve as semantic anchors that describe the desired categories in a way that aligns with typical real estate photography. Here we demonstrate four example pseudo-prompts:

\begin{itemize}
    \item \textit{indoor:} a bright indoor photo of a living room, bedroom, or kitchen in a residential house
    \item \textit{outdoor:} an outdoor photo of a house backyard, garden, facade, or swimming pool
    \item \textit{mixed:} a property photo showing both indoor rooms and outdoor views through large windows or doors
    \item \textit{floorplan:} a 2D floorplan diagram or architectural layout of a house with room labels
\end{itemize}

These pseudo-prompts are intentionally descriptive and domain-specific. They include visual cues like ``bright" for indoor appeal, ``large windows or doors" for mixed transitional views, ``room labels" for floorplans. The prompts were encoded into the same latent space as the image embeddings by passing through CLIP's text encoder, producing four text embeddings that represent conceptual anchors for the categories.

Each image's CLIP embedding was compared to these four text embeddings using cosine similarity:
\begin{equation}
\label{cos:sim}
\text{sim}(i,t_m) = \cos(\theta_{i,m}) = \frac{i \cdot t_m}{\|i\| \|t_m\|}
\end{equation}
where $i$ is the normalized image embedding and $t_m$ is the normalized embedding of prompt $m$. The cluster is mapped to the prompt with the highest similarity score.

The pseudo-prompts act as semantic anchors, i.e., the contrastive loss during pre-training ensures that images visually similar to the described concepts are close to the corresponding text embeddings in the joint space. By comparing each image embedding clusters to the prompt embeddings, we classified images into categories based on learned semantic similarity, without any manual labeling or supervised fine-tuning.

\subsubsection{Visual Feature Construction}

The feature construction process combined three complementary components: reduced embedding dimensions, category presence indicators, and category-conditional interactions.

The first component consists of low-dimensional embeddings obtained from UMAP. Each dimension captured latent visual patterns, including overall composition, color palette, object presence, lighting conditions, and stylistic elements.

The second component comprises four binary indicators, one for each semantic category: indoor, outdoor, mixed, and floorplan. Each indicator was set to 1 if at least one image for the property was assigned to the corresponding category, and 0 otherwise. For example, the indoor indicator reflects whether interior photos are available, which often correlates with assessments of livability and condition; the outdoor indicator signals curb appeal, garden, or facade quality; the mixed indicator captures transitional or lifestyle-oriented views; and the floorplan indicator denotes the architecture of layout diagrams, which enhance buyer confidence in spatial functionality.

The third component consists of interaction terms that modulate selected dimensions from the reduced embedding space by category presence. Specifically, we multiplied one or more embedding dimensions by the corresponding category indicator. They allowed the model to learn that certain visual patterns (captured by the embedding dimensions) have amplified or diminished importance depending on the category. For example, a dimension representing outdoor lighting or greenery contributes more strongly to price when the outdoor category is present. This interaction mechanism captures non-linear, modality-specific influences that would be difficult to learn from additive features alone.

This feature construction step completes the visual processing branch by transforming raw perceptual input into compact, semantically meaningful, and price-relevant signals. The reduced embeddings retain core visual patterns, the binary indicators provide explicit category presence, and the interaction terms introduce conditional weighting; all without requiring labeled supervision.

\subsection{Feature Fusion and Price Prediction}

After independently processing the three modalities (structured, textual, and visual), the extracted features were fused into a unified representation. This fusion strategy allows the model to leverage complementary information from numerical attributes, rich contextual signals from text, and semantic visual cues from property images. The final feature vector was formed by concatenating the processed features from all three sources. Categorical features were encoded to numerical representations, and numerical features were scaled using a robust transformation that centers on the median and normalizes by interquartile range.

To optimize the model performance, hyperparameters were tuned systematically using Optuna \citep{akiba2019optuna}. The final regressor model was trained using the optimal hyperparameters to predict prices.

%% 4. Experiments

\section{Experiments}

We conduct a thorough experimental evaluation of our framework on a real estate dataset from Australia. The dataset covers sold properties from a range of Australian locations. To validate its effectiveness, we benchmark our model against established baselines and state-of-the-art models for house price prediction.

Our approach achieves state-of-the-art performance, with substantial reductions in both mean absolute error (MAE) and RMSE. An ablation study further demonstrates the critical contribution of key feature extraction from textual data and image categorization in improving the accuracy of house price prediction.

\subsection{Dataset Composition}
\label{subsec:dataset_composition}

The experiments were conducted using the house transaction records collected from a leading real estate platform\footnote{\url{https://www.realestate.com.au/}}. The dataset comprises 52,917 transactions spanning 2021 to 2023. It also incorporates details on nearby points of interest, including schools, shops, and train stations. Each record features three complementary modalities: 29 structured attributes such as number of bedrooms, bathrooms, parking, property type, distance to transport hubs, latitude, longitude, median rent, income levels, school rankings, and various facility indicators; textual attributes consisting of a property heading and content description that offer insights into condition, location advantages, seller intent; and a total of 264,255 uncategorized property images capturing diverse visual aspects and views of the houses. The dataset for this experiment is available on Zenodo \citep{hossain2026dataset}.

\subsection{Results}

We evaluate our model's performance through a comprehensive comparative analysis and detailed ablation studies. The comparative analysis benchmarks our full multimodal framework against established baselines that employ varying combinations of structured, textual, and visual data. The ablation studies systematically isolate the contribution of each modality and design choice. These results highlight how the integration of spatio-temporal and contextual textual features, combined with categorized visual embeddings, drives substantial improvements in prediction accuracy over traditional and prior multimodal approaches.

\subsubsection{Performance Metrics}
\label{sec:performance_metrics}

The normalized RMSE and MAE are used to quantify the deviation between the predicted house prices and the actual sale prices, with lower values indicating better predictive accuracy.

\textbf{Root Mean Squared Error (RMSE).}
The RMSE places greater emphasis on larger errors by squaring the differences before averaging and then taking the square root. It is computed as:

\begin{equation}
\text{RMSE} = \sqrt{\frac{1}{N} \sum_{i=1}^{N} (y_i - \hat{y}_i)^2}
\end{equation}

where $N$ is the number of samples, $y_i$ is the actual sale price of the $i$-th property, and $\hat{y}_i$ is the predicted price.

\textbf{Mean Absolute Error (MAE).}
The MAE calculates the average absolute difference between the predicted and true values. It is defined as:

\begin{equation}
\text{MAE} = \frac{1}{N} \sum_{i=1}^{N} |y_i - \hat{y}_i|
\end{equation}

where the symbols denote the same meanings as in RMSE.

\subsubsection{Inference Settings}

The final dataset was randomly partitioned into training, validation, and test sets using a 70\%/15\%/15\% split. The validation set was used for early stopping and hyperparameter tuning via Optuna with 5-fold cross-validation, and the test set was used for final performance evaluation. During the inference phase, the trained LightGBM model was used to predict house prices on test data. Table \ref{tab:inference_settings} summarizes the key inference settings and configurations used for each component of our multimodal pipeline.

\begin{table}[htbp]
\caption{Inference Settings}
\label{tab:inference_settings}
\begin{tabular}{@{}>{\raggedright\arraybackslash}p{0.25\textwidth}>{\raggedright\arraybackslash}p{0.70\textwidth}@{}}
\hline
\textbf{Component} & \textbf{Details} \\
\hline
Hardware & NVIDIA Tesla V100-PCIE-32GB GPU, Intel Xeon Silver 4214R CPU, 383 GB RAM \\
\hline
Batch Size & 32 \\
\hline
Training Hyperparams & - LightGBM (\textit{n\_estimators = 950, max\_depth = 12, learning\_rate = 0.04, num\_leaves = 128, min\_child\_samples = 20, subsample = 0.85, colsample\_bytree = 0.8, reg\_lambda = 1.5}) \\
 & - Optuna (\textit{trials = 150, 5-fold cross-validation}) \\
 & - UMAP (\textit{n\_components = 10, n\_neighbors = 15}) \\
\hline
Structured Features & Numerical and categorical features (\textit{including engineered polynomials, interaction features, and locality statistics}) \\
\hline
Textual Features & Most frequently occurring spatio-temporal and contextual features in the dataset (\textit{proximity\_minutes, proximity\_km, near\_shops, near\_transport, near\_schools, is\_new, is\_quiet, is\_busy, is\_central, has\_view, north\_facing, urgency\_level, is\_recently\_renovated, is\_old, has\_luxury\_finishes, sale\_season}) \\
\hline
Visual Features & Reduced UMAP embeddings, category indicators (\textit{indoor, outdoor, mixed, floorplan}), and category-conditional interaction terms \\
\hline
LLM Configuration & LLaMA-2-7B \\
\hline
VLM Configuration & CLIP ViT-B/32 \\
\hline
Runtime Configuration & GPU acceleration for image embedding and LLM inference \\
\hline
\end{tabular}
\end{table}

\subsubsection{Computation and Memory Overhead}
The proposed framework incurs moderate computational and memory overhead due to the multimodal feature extraction stages. Textual feature extraction with LLaMA-2-7B requires approximately 12 GB of GPU memory and takes roughly 1.5 to 2 seconds per property description on NVIDIA Tesla V100-PCIE-32GB GPU. Visual processing with CLIP ViT-B/32 is significantly lighter, where embedding generation runs at approximately 55 to 60 images per second, with peak GPU memory usage below 4 GB. UMAP dimensionality reduction and subsequent clustering add negligible cost at inference time.

End-to-end feature extraction for the full dataset required about 18 hours, which was primarily dominated by LLM inference. Once features are extracted, training the final LightGBM model typically completes in under 3 to 3.5 minutes. At inference time, a single property can be processed in under 5 seconds when both LLM and VLM features are generated on the fly. Overall, the framework offers a practical trade-off between predictive performance and computational cost, remaining feasible for both research and real-world deployment scenarios.

\subsubsection{Comparative Analysis}

We compare our method against several representative baselines, each utilizing different feature sets. Model performance is assessed using RMSE and MAE. Traditional hedonic regression relies exclusively on structured attributes. Geospatial models (e.g., Spatial Autoregressive (SAR) Estimator) incorporate structured data along with spatio-temporal features. Graph-based methods (e.g., GSNE) integrate structured and geospatial signals. Multimodal deep learning frameworks (e.g., MHPP, the multimodal deep learning based house price prediction approach of \citet{hasan2024multi}) combine structured data with both textual and uncategorized visual features. In contrast, our method uniquely leverages structured attributes, spatio-temporal and contextual features extracted from text via large language models, and categorized visual embeddings obtained through unsupervised vision-language alignment.

Table \ref{tab:comparative} presents the comparative results. We evaluated the same full multimodal feature set of our model across several widely used regressors: Lasso, Elastic Net (E-Net), Kernel Ridge Regression (KRR), XGBoost (XGB), and LightGBM (LGBM). Compared to the strongest baseline, our model reduces RMSE by 21.9\% (from 0.160 to 0.125) and MAE by 29.4\% (from 0.112 to 0.079). These improvements demonstrate the effectiveness of our proposed framework, particularly the LLM-based structured extraction of spatio-temporal and contextual features, as well as the unsupervised categorization of images into semantically meaningful classes.

\begin{table}[htbp]
\caption{Comparative Performance Analysis}
\label{tab:comparative}
\begin{tabular}{lcc}
\hline
Method & RMSE & MAE \\
\hline
Hedonic Regression & 0.294 & 0.217 \\
Spatial Autoregressive Estimator (SAR) & 0.201 & 0.152 \\
Graph-based (GSNE) & 0.183 & 0.128 \\
Multimodal DL (MHPP) & 0.160 & 0.112 \\
\hline
\multirow{5}{*}{Ours\hspace{0.4em}\vrule} \hspace{0.1em} Lasso   & 0.196 & 0.127 \\
                    \hspace{3em} E-Net  & 0.180 & 0.119 \\
                    \hspace{3em} KRR    & 0.138 & 0.089 \\
                    \hspace{3em} XGB    & 0.131 & 0.085 \\
                    \hspace{3em} LGBM   & \textbf{0.125} & \textbf{0.079} \\
\hline
\end{tabular}
\end{table}

Table \ref{tab:comparative} also demonstrates that simpler linear models (Lasso and E-Net) show limited capability in capturing complex interactions, while tree-based ensemble models (XGB and LGBM) benefit substantially from our rich multimodal features.

\subsubsection{Ablation Studies}

To quantify the contribution of each modality, we perform ablation studies by progressively omitting components of our framework. The results are summarized in Table~\ref{tab:ablation_results}, based on the five representative regressors. LGBM consistently achieves the best performance across all variants, where the full multimodal model with visual categorization yields the lowest errors (RMSE 0.125, MAE 0.079).

\begin{table}[htbp]
\caption{Performance Comparison of Different Components}
\label{tab:ablation_results}
\begin{tabular}{l>{\raggedright\arraybackslash}p{4.6cm}ccccc}
\hline
Metric & Modality & Lasso & E-Net & KRR & XGB & LGBM \\
\hline
\multirow{7}{*}{RMSE\hspace{0.4em}\vrule}
& Structured & 0.362 & 0.347 & 0.305 & 0.299 & 0.294 \\
& Structured + Textual & 0.230 & 0.211 & 0.170 & 0.165 & 0.158 \\
& Structured + Visual & 0.228 & 0.209 & 0.169 & 0.161 & 0.156 \\
& Structured + Textual + Visual\newline(no categorization) & 0.214 & 0.194 & 0.156 & 0.145 & 0.140 \\
& Structured + Textual + Visual\newline(with categorization) & 0.196 & 0.180 & 0.138 & 0.131 & \textbf{0.125} \\
\hline
\multirow{7}{*}{MAE\hspace{0.6em}\vrule}
& Structured & 0.266 & 0.256 & 0.228 & 0.224 & 0.217 \\
& Structured + Textual & 0.149 & 0.140 & 0.119 & 0.114 & 0.110 \\
& Structured + Visual & 0.148 & 0.138 & 0.109 & 0.101 & 0.097 \\
& Structured + Textual + Visual\newline(no categorization) & 0.131 & 0.127 & 0.099 & 0.094 & 0.088 \\
& Structured + Textual + Visual\newline(with categorization) & 0.127 & 0.119 & 0.089 & 0.085 & \textbf{0.079} \\
\hline
\end{tabular}
\end{table}

The different ablation results summarize that, the Structured + Textual variant serves as the strongest baseline without visual information, achieving an RMSE of 0.158 and MAE of 0.110 with LightGBM. Replacing textual with visual features (Structured + Visual) yields comparable performance (RMSE 0.156, MAE 0.097), indicating that both modalities provide substantial value on their own, with visual features slightly outperforming textual in this dataset. Combining both modalities without categorization (Structured + Textual + Visual no categorization) further reduces errors to RMSE 0.140 and MAE 0.088, confirming the complementary nature of textual context and raw visual patterns.

The full model, which additionally applies semantic categorization to the visual modality, delivers the best performance with RMSE 0.125 and MAE 0.079. This represents a meaningful improvement over the uncategorized visual case. The gain arises from the unsupervised grouping of images into four semantically meaningful classes which enables explicit modeling of buyer-relevant visual aspects and category-conditional effects through interaction terms.

\textbf{Impact of Textual Features.} Incorporating spatio-temporal and contextual textual features extracted via large language model substantially enhances prediction accuracy over structured data alone. It reflects the ability of the model to capture latent signals that are either absent or only implicitly present in numerical attributes.

\textbf{Impact of Visual Features and Categorization.} Visual features contribute substantially even in their uncategorized form, reducing RMSE from 0.158 (structured + textual) to 0.140 when added without categorization. This improvement demonstrates that raw CLIP embeddings, after UMAP reduction, encode useful perceptual signals related to property aesthetics. However, applying semantic categorization yields a further reduction to RMSE 0.125 and MAE 0.079. This additional gain confirms that grouping images into domain-relevant classes enables the model to learn more interpretable visual representations. The categorization step bridges low-level visual patterns with high-level buyer priorities, making visual information more actionable for price prediction.

\label{subsec:ablation_model_framework}
\textbf{Impact of Different Model Frameworks.} To examine the sensitivity of our framework to the choice of foundation models, we evaluated several combinations of large language models for textual feature extraction, and vision-language models for visual categorization. All experiments retained the same unsupervised image categorization pipeline of quality filtering, embedding generation, clustering, and matching with pseudo-prompts to assign semantic labels to the images. All used the downstream LightGBM as the final regressor. The results are reported in Table~\ref{tab:llm_vlm}.

The primary proposed configuration is LLaMA-2-7B + CLIP-ViT. In this setting, both the textual extraction and the visual embedding and categorization steps rely on open-source models. For comparison, we replaced the text language model with stronger proprietary alternatives like GPT-5.6 Luna, Claude Opus 4.6, and Gemini-2.5-Pro, while keeping the CLIP-ViT image embeddings and the remainder of the unsupervised visual pipeline unchanged. Because the standard text embedding models of Luna and Claude do not currently provide vector embeddings from raw images, the experiments with them reuse the same CLIP-ViT visual representations (Section \ref{subsec:visual_embedding_generation}). Only the textual feature extraction is performed by the respective proprietary model.

In one additional configuration, we replaced CLIP with Gemini Embedding 2 for the visual embeddings generation with the same unsupervised categorization pipeline, and allowing Gemini-2.5-Pro to handle feature extraction from textual descriptions along with it.

\begin{table}[htbp]
\caption{Performance on Different Model Frameworks}
\label{tab:llm_vlm}
\begin{tabular}{llcc}
\hline
\textbf{LLM} & \textbf{VLM} & \textbf{RMSE} & \textbf{MAE} \\
\hline
LLaMA-2-7B & CLIP-ViT & 0.125 & 0.079 \\
Claude Opus 4.6 & CLIP-ViT & 0.132 & 0.083 \\
GPT-5.6 Luna & CLIP-ViT & 0.130 & 0.082 \\
Gemini-2.5-Pro & CLIP-ViT & 0.127 & 0.079 \\
Gemini-2.5-Pro & Gemini Embedding 2 & \textbf{0.124} & \textbf{0.077} \\
\hline
\end{tabular}
\end{table}

Table~\ref{tab:llm_vlm} shows that performance differences remain relatively small. The fully Gemini-based configuration attains the lowest errors (RMSE 0.124, MAE 0.077). The open-source LLaMA-2-7B + CLIP-ViT baseline is highly competitive, matching the best MAE among the CLIP-based variants and trailing the strongest result by only 0.001 in RMSE. Configurations that pair proprietary LLMs with CLIP-ViT yield comparable or slightly higher errors, confirming that the gains from stronger language models are modest once the visual pipeline is held fixed.

These findings highlight two important insights. First, the core contribution of this work lies in the proposed methodology, namely the generalized prompt-based extraction of spatio-temporal and contextual features from text combined with the unsupervised pseudo-prompt categorization of property images, rather than in the specific choice of foundation model. Second, the framework is robust. It delivers strong and consistent performance even with fully open-source and locally deployable models, while still allowing slight gains when more powerful commercial models are available. This robustness enhances the practical value of the approach, as practitioners can select models according to their computational budget, privacy requirements, and deployment constraints.

\subsubsection{Qualitative Analysis}

To complement the quantitative evaluation, we perform a detailed qualitative analysis of our feature extraction pipeline. This analysis highlights the strengths and limitations of the LLM for textual feature extraction and the VLM for visual categorization.

\textbf{Textual Feature Extraction (LLM).} The feature extraction was primarily performed using LLaMA-2-7B under the generalized prompt. We additionally evaluated stronger closed-source models (Section \ref{subsec:ablation_model_framework}) on the same 200 randomly selected samples for qualitative comparison.

LLaMA-2-7B produced coherent and useful feature sets for 162 out of 200 samples. Representative successes included:
\begin{itemize}
    \item \textit{``Only short 5 min. drive to Manor Lakes"} correctly extracted as
    \{\textit{proximity\_minutes}: 5, \textit{near\_lakes}: 1\}.
    \item \textit{``Urgent Sale! All offers Invited! Call today!"} correctly extracted as
    \{\textit{urgency\_level}: 2, \textit{seller\_intent\_urgent}: 1\}.
\end{itemize}

GPT 5.6 Luna showed the highest consistency in feature naming and better handling of vague promotional languages like ``idyllic lifestyle within easy reach of all the district", achieving reasonable features extractions on 189 out of the 200 samples. However, overly optimistic narratives like ``redefine the meaning of carefree living" often resulted in no useful features.

\textbf{Visual Categorization (VLM).} The VLM-based categorization was evaluated on all 283 available images associated with the 200 houses using CLIP. It succeeded in correctly classifying 266 out of 283 images verified through manual observation. The model could reliably assign certain clear cases:
\begin{itemize}
    \item Standalone backyard or facade photos to the \textit{outdoor} category.
    \item Clear interior living rooms or kitchens to the \textit{indoor} category.
    \item Explicit 2D floorplan diagrams to the \textit{floorplan} category.
\end{itemize}

Misclassifications occurred mostly in cases where the image content only partially or weakly matched the intended prompts:
\begin{itemize}
    \item Transitional scenes like indoor rooms with large open windows revealing outdoor views were sometimes classified as \textit{indoor} instead of \textit{mixed}, when the outdoor portion was not visually dominant but there was a moderately large outdoor portion in the photo.
    \item Facade photos containing subtle indoor elements like a clearly visible door frame were occasionally grouped as \textit{mixed} rather than \textit{outdoor}.
\end{itemize}

These qualitative results demonstrate that our LLM and VLM pipelines are highly effective for the majority of real-world property listings. The extracted features capture critical buyer-oriented signals that are typically missed by structured-only baselines.

\subsubsection{Model Interpretability with SHAP}

To gain deeper insights into which features most strongly influence the property price predictions, we apply SHAP \citep{louhichi2023shapley} analysis on the final model. We focus the interpretability analysis here on our model because the existing multimodal models do not perform any semantic categorization of visual images nor structured spatio-temporal feature extraction from textual descriptions. As a result, their raw high-dimensional embeddings offer limited interpretability compared to our semantically meaningful features.

\begin{figure}[htbp]
\centering
\includegraphics[width=0.7\textwidth]{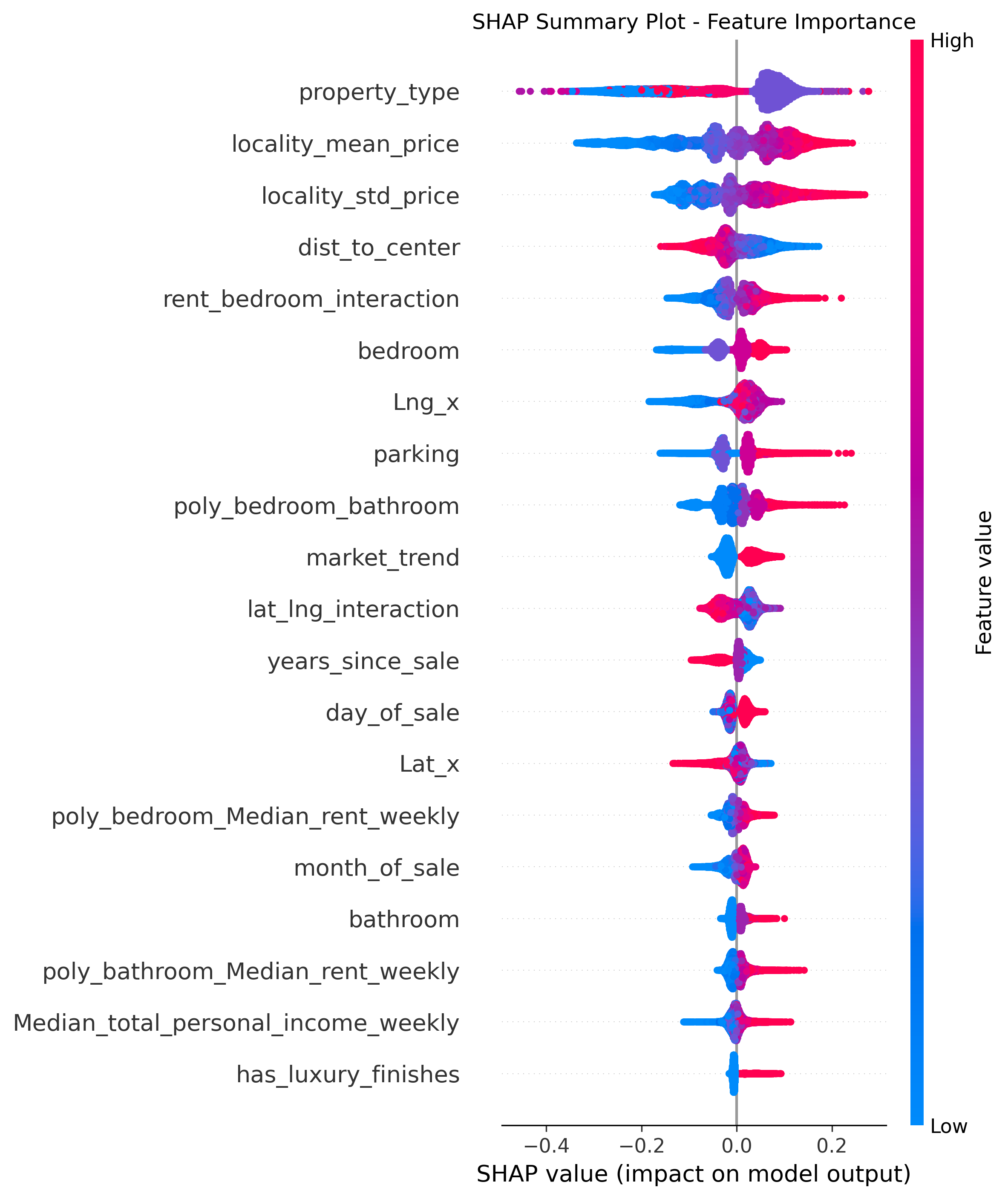}
\caption{SHAP Summary Plot for Structured Features Showing Feature Importance}
\label{fig:shap_summary}
\end{figure}

Figure~\ref{fig:shap_summary} shows the SHAP summary plot for the structured features, illustrating the global importance and impact direction of the top features. These results confirm that property-type information, location and proximity-related features most influence the predictions.

\begin{figure}[htbp]
\centering
\includegraphics[width=0.7\textwidth]{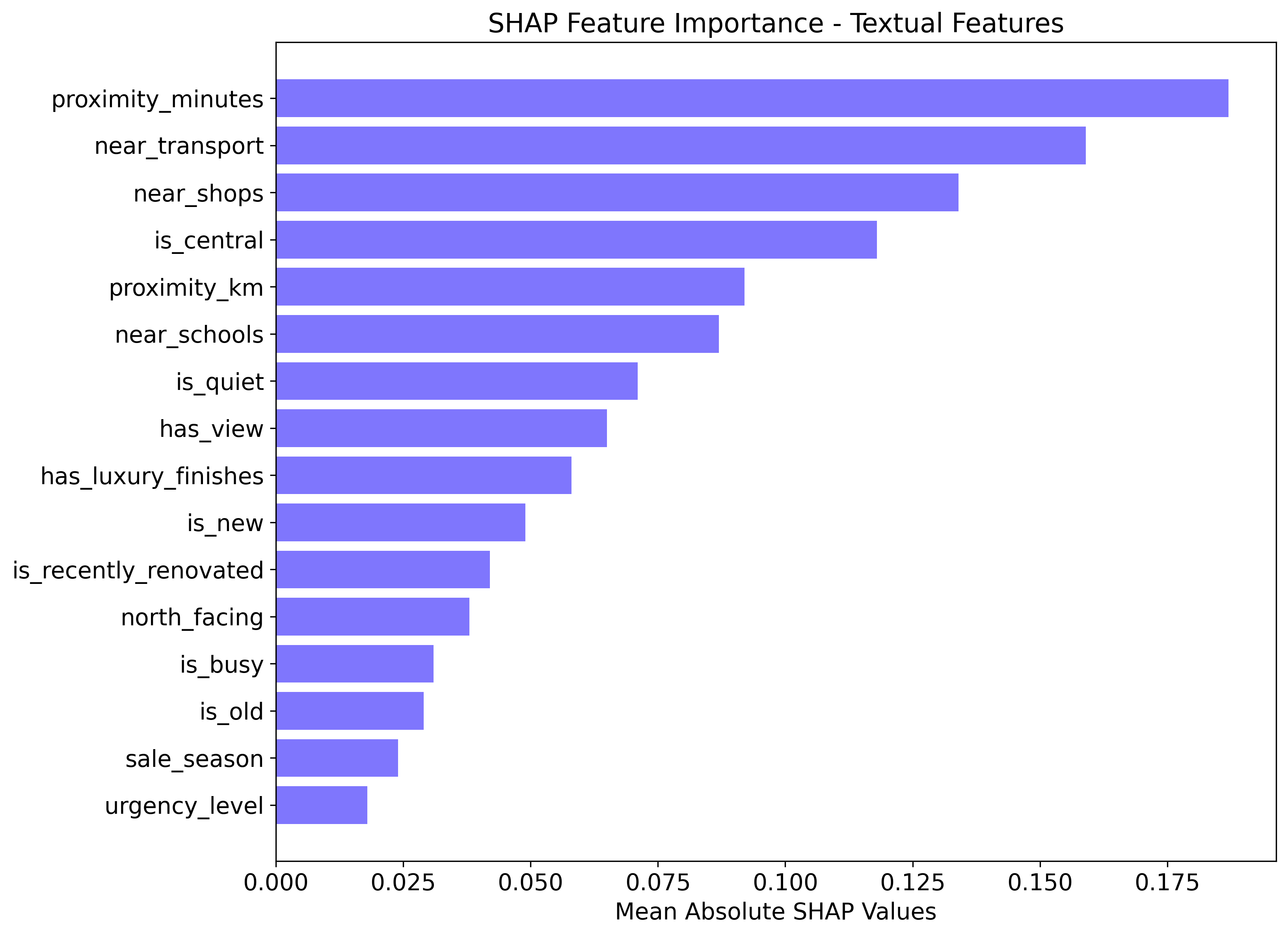}
\caption{SHAP Feature Importance Plot for Textual Features}
\label{fig:shap_textual_features}
\end{figure}

Figure~\ref{fig:shap_textual_features} presents the SHAP feature importance for the spatio-temporal and contextual features extracted by LLaMA-2-7b. These results show that proximity to amenities, renovation status are among the strongest textual drivers of property price in our model.

\begin{figure}[htbp]
\centering
\includegraphics[width=0.7\textwidth]{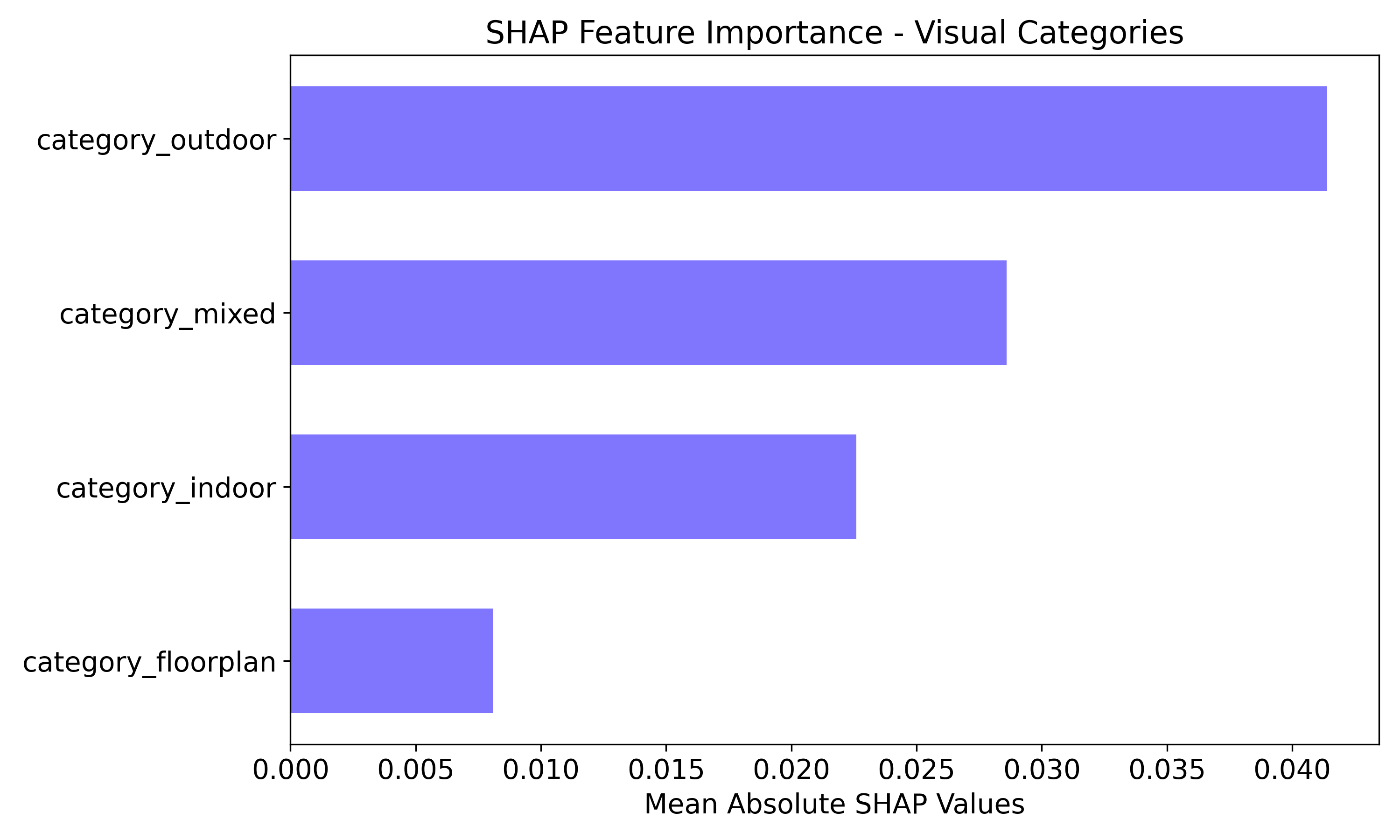}
\caption{SHAP Feature Importance Plot for Visual Categories}
\label{fig:shap_visual_importance}
\end{figure}

The SHAP analysis also quantifies the contribution of the four visual category indicators. Figure~\ref{fig:shap_visual_importance} shows the mean absolute SHAP values for each category, indicating their impact on the predicted house price. The \textit{outdoor} category has the highest average impact, followed by \textit{mixed}. The \textit{indoor} category also contributes meaningfully after \textit{mixed} category. Lastly, the \textit{floorplan} category has a smaller but positive effect.

Overall, the SHAP analysis demonstrates that our model learns intuitive and economically meaningful patterns. The high importance of engineered location and interaction features, along with the contributions from textual and categorized visual features, provides strong evidence of the effectiveness and interpretability of our multimodal framework.

%% 5. Conclusion

\section{Conclusion}

In this paper, we presented a novel multimodal framework for house price prediction that integrates structured attributes, textual descriptions, and visual images to achieve superior accuracy and interpretability. By leveraging large language models for extracting spatio-temporal and contextual features from unstructured text, and an unsupervised vision-language approach for categorizing property images into semantically meaningful classes without manual labeling, our method addresses key limitations of traditional and prior multimodal techniques.

Our framework has practical implications for real estate stakeholders, buyers, sellers, and governments, offering a scalable and interpretable tool for fair transactions, automated market analysis, and informed decision-making. Future work could explore incorporating additional modalities like satellite imagery, or adapting to diverse geographic regions for broader applicability.

%% Back matter

\backmatter

\section*{Statements and Declarations}

\bmhead{Funding}
No funding was received for conducting this study.

\bmhead{Competing interests}
The authors have no competing interests to declare that are relevant to the content of this article.

\bmhead{Data availability}
The dataset analyzed in this study is available on Zenodo \citep{hossain2026dataset} at \url{https://doi.org/10.5281/zenodo.23757153}.

\bmhead{Code availability}
The complete code for this paper is available on GitHub at \url{https://github.com/rh2975/LLM-VLM-HousePrice.git}.

%% Same statement must also be entered in the submission system.
\bmhead{Author contributions}
Conceptualization: Khandokar Md. Rahat Hossain, Souvik Das; Methodology: Khandokar Md. Rahat Hossain, Souvik Das; Software: Khandokar Md. Rahat Hossain, Souvik Das; Data curation: Khandokar Md. Rahat Hossain, Souvik Das; Formal analysis: Khandokar Md. Rahat Hossain, Tahsina Hashem; Visualization: Khandokar Md. Rahat Hossain; Resources: Tanzima Hashem; Writing -- original draft: Khandokar Md. Rahat Hossain, Souvik Das; Writing -- review and editing: all authors; Supervision: Tahsina Hashem, Mohammed Eunus Ali, Tanzima Hashem. All authors read and approved the final manuscript.

%%=====================================================================%%
%% Appendices
%%=====================================================================%%

% \begin{appendices}

% \end{appendices}

%%=====================================================================%%
%% References
%% (style is set by the class option; do not add \bibliographystyle)
%%=====================================================================%%

%% The reference list is embedded below (BibTeX output of refs.bib with
%% sn-basic.bst, author-year), as recommended by the Springer Nature LaTeX
%% author support guide. It contains only works cited in the text, as the
%% journal requires. To regenerate it after editing refs.bib, temporarily
%% replace the thebibliography block with \bibliography{refs}, run
%% pdflatex, bibtex, pdflatex, then paste the new main.bbl back here.

\end{document}